\documentclass{article}
\usepackage{amssymb}
\usepackage{authblk}
\usepackage{graphicx}
\usepackage{multicol}
\usepackage{amsmath,amsfonts}
\usepackage{algorithm}
\usepackage{algpseudocode}
\usepackage{algorithm}
\usepackage{caption}
\usepackage{subfigure}
\usepackage{subcaption}
\usepackage{enumitem}
\setlist[enumerate]{label=(\roman*),
                    itemindent=3em,
                    leftmargin=2em, 
                    }

\usepackage[hidelinks]{hyperref}

\usepackage{bbm}
\hypersetup{
    colorlinks=true,
    linkcolor=purple,
    filecolor=purple,      
    urlcolor=purple,
    citecolor=purple,
}

\usepackage{titlesec}
\usepackage{chngcntr}
\counterwithin{paragraph}{section}

\titleformat{\paragraph}[runin]
  {\normalfont\bfseries}{}{0pt}{}[.]
\titlespacing*{\paragraph}{0pt}{1ex}{1em}

\usepackage{stfloats}
\usepackage{booktabs}   
\usepackage{siunitx}    
\usepackage{wrapfig}    
\usepackage[most]{tcolorbox}
\usepackage{tikz}
\newcommand{\stepicon}[1]{%
    \tikz[baseline=-0.5ex] \node[draw,circle,inner sep=1pt,fill=blue!20,font=\footnotesize\sffamily]{#1};%
}
\usepackage{enumitem}

\usepackage{xcolor}
\usepackage{tcolorbox}
\newenvironment{tightlist}%
{\begin{list}{$\bullet$}{%
    \setlength{\topsep}{0in}
    \setlength{\partopsep}{0in}
    \setlength{\itemsep}{0in}
    \setlength{\parsep}{0in}
    \setlength{\leftmargin}{1.5em}
    \setlength{\rightmargin}{0in}
}
}%
{\end{list}
}
\usepackage{color}
\usepackage{amssymb} 

\usepackage{setspace}
\usepackage{mathtools}

\usepackage[final]{corl_2025} 
\title{PrioriTouch: Adapting to User Contact Preferences for Whole-Arm Physical Human-Robot Interaction}

\author{
  Rishabh Madan$^{1}$, Jiawei Lin$^{1}$, Mahika Goel$^{1}$, Amber Li$^{1}$, Angchen Xie$^{1}$, \\Xiaoyu Liang$^{1}$, Marcus Lee$^{1}$, Justin Guo$^{1}$, Pranav N. Thakkar$^{1}$, Rohan Banerjee$^{1}$, \\Jose Barreiros$^{2}$, Kate Tsui$^{2}$, Tom Silver$^{1}$, Tapomayukh Bhattacharjee$^{1}$\\
  $^{1}$Cornell University, $^{2}$Toyota Research Institute \\
}

\begin{document}
\maketitle

\begin{center}
    \captionsetup{type=figure}
    \includegraphics[width=\textwidth]{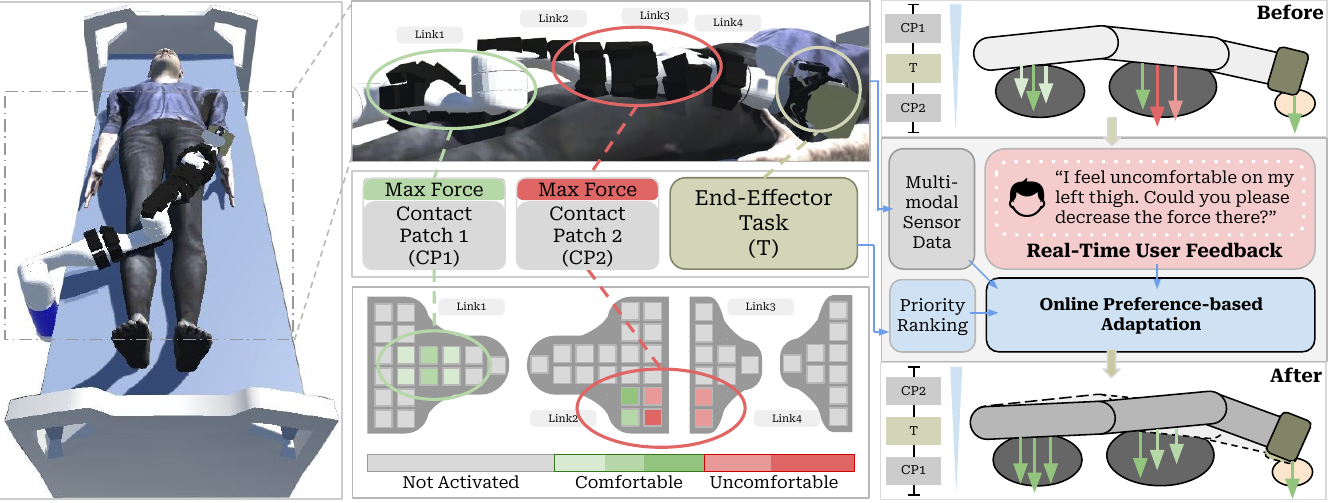}
    \caption{Overview of PrioriTouch: A hierarchical controller coupled with simulation‑in‑the‑loop online contact preference learning aids in personalizing multi-contact physical human-robot interaction for safe and comfortable whole-arm manipulation.\looseness=-1}
    \label{fig:first_fig}
\end{center}

\begingroup
\hyphenpenalty=10000
\exhyphenpenalty=10000
\begin{abstract}
    Physical human–robot interaction (pHRI) requires robots to adapt to individual contact preferences, such as where and how much force is applied. Identifying preferences is difficult for a single contact; with whole-arm interaction involving multiple simultaneous contacts between the robot and human, the challenge is greater because different body parts can impose incompatible force requirements. In caregiving tasks, where contact is frequent and varied, such conflicts are unavoidable. With multiple preferences across multiple contacts, no single solution can satisfy all objectives--trade-offs are inherent, making prioritization essential. We present PrioriTouch, a framework for ranking and executing control objectives across multiple contacts. PrioriTouch can prioritize from a general collection of controllers, making it applicable not only to caregiving scenarios such as bed bathing and dressing but also to broader multi-contact settings. Our method combines a novel learning-to-rank approach with hierarchical operational space control, leveraging simulation-in-the-loop rollouts for data-efficient and safe exploration. We conduct a user study on physical assistance preferences, derive personalized comfort thresholds, and incorporate them into PrioriTouch. We evaluate PrioriTouch through extensive simulation and real-world experiments, demonstrating its ability to adapt to user contact preferences, maintain task performance, and enhance safety and comfort. Website: \href{https://emprise.cs.cornell.edu/prioritouch/}{https://emprise.cs.cornell.edu/prioritouch}.
\end{abstract}
\endgroup

\newcommand{\prefpath}{\mathcal{P}_{\text{path}}}      
\newcommand{\pwp}[1]{\mathbf{p}(#1)}                  
\newcommand{\pcurr}[1]{\mathbf{p}^{\mathrm{curr}}_{#1}}
\newcommand{\pdes}[1]{\mathbf{p}^{\mathrm{des}}_{#1}}  
\newcommand{\wpidx}[1]{i_{#1}}                         

\newcommand{\skel}{\mathcal{B}}                        
\newcommand{\hpose}{\mathbf{h}_t}                      

\newcommand{\contactset}[1]{\mathcal{C}_{#1}}          
\newcommand{\contactseton}[2]{\mathcal{C}_{#1}(#2)}    
\newcommand{\contactforce}[2]{\mathbf{f}_{#1}(#2)}     
\newcommand{\bodyforce}[2]{f_{#1}(#2)}                 

\newcommand{\state}{\mathsf{s}_t}                      

\newcommand{\usermodel}{\mathcal{H}}                   
\newcommand{\comfort}[1]{f^{\max}(#1)}                

\newcommand{\feedback}[1]{\phi_{#1}}                   
\newcommand{\feedbackpair}[2]{(#1,#2)}                 

\newcommand{\Jpose}[1]{J_{\text{pose}}(#1)}            
\newcommand{\Jforce}[2]{J_{\text{force},#2}(#1)}       
\newcommand{\Jset}[1]{\mathcal{J}_{#1}}                

\newcommand{\pisig}{\pi_\sigma}                        
\newcommand{\Ascr}[1]{\mathcal{A}_{#1}}                

\newcommand{\ctx}[1]{\mathbf{z}_{#1}}                  
\newcommand{\rew}[1]{r_{#1}}                           

\newcommand{\Kp}{\mathbf{K}_p}
\newcommand{\Kd}{\mathbf{K}_d}
\newcommand{\Kf}{\mathbf{K}_f}

\section{Introduction}
\vspace{-5pt}

Physical human–robot interaction (pHRI) requires physical contact. Contact is not uniform: individuals have distinct preferences for acceptable forces and contact locations \cite{hollinger1993factors, o2017perceptions, de2022permission, chen2011touched, behrens2022statistical}. For pHRI to be safe and effective, robots must personalize their behavior, and a critical aspect of personalization is contact preferences. Even for a single contact, identifying and respecting these preferences while ensuring task success is challenging. Many physical robot caregiving tasks, such as bathing \cite{king2010towards, madan2024rabbit}, dressing \cite{zhang2017personalized, kapusta2019personalized}, and transferring \cite{10974069}, require whole‑arm pHRI \cite{salisbury1987whole}, where multiple segments of the robot arm simultaneously touch the human body. For example, during bed bathing (Fig.~\ref{fig:first_fig}), the robot may need to reach over a user to wipe the upper arm while maintaining comfortable forces on the torso and shoulder. Although whole‑arm manipulation expands workspace and improves maneuverability, it also exacerbates conflicts: different body parts can impose incompatible force requirements, and no single policy can satisfy all objectives.

To bootstrap personalization, we elicit population‑level contact preferences offline and use them to seed a conservative base policy. However, a one‑size‑fits‑all policy is insufficient: (i) stated preferences can diverge from realized comfort under true contact (pressure/shear, approach, speed, duration); and (ii) preferences are context‑dependent and time‑varying (posture, clothing, fatigue). Therefore, online interaction is necessary to accommodate individual preferences. Experimenting directly with the user is risky and inefficient because each update can involve repeated physical contact and multiple feedback exchanges. This increases the user's cognitive workload, prolongs the interaction, and may cause discomfort when forces are suboptimal or excessive.

We introduce PrioriTouch, a framework that casts contact preference learning as a learning‑to‑rank problem over control objectives. Given a reference trajectory produced by a high‑level policy (e.g., a contact‑aware planner generating end‑effector or joint‑space paths), PrioriTouch instantiates pose‑tracking and force‑regulation objectives from the current contact state. We develop LinUCB‑Rank, a contextual bandit that learns a priority policy; H‑OSC \cite{sentis2004prioritized} then executes this ordering as a null space hierarchy, translating high‑level preferences into low‑level control. We initialize the policy with conservative priors derived from population-level user-study statistics. During interaction, LinUCB-Rank adapts the ordering online using sparse user feedback while safely refining the policy via simulation-in-the-loop learning before deploying it in real-world interactions. The framework is controller-agnostic: it can rank heterogeneous objectives, enabling principled trade-offs across simultaneous objectives. \looseness=-1

We evaluate PrioriTouch across simulated and real-world environments, progressively increasing in complexity and realism. First, we design a simplified simulation scenario with predefined contacts and a static end-effector pose to isolate and specifically assess LinUCB-Rank's ability to learn user contact preferences. Second, we demonstrate PrioriTouch in a simulated caregiving scenario involving robot-assisted bed bathing, requiring whole-arm contact to safely wipe a user’s limbs. Third, we showcase our approach’s capability in intricate multi-contact scenarios through a real-world 3D goal-reaching maze with multiple vertical cylinders representing distinct body-part contact preferences. Finally, we validate PrioriTouch’s practical feasibility by performing a realistic caregiving task in a user study with human subjects.

Our contributions are summarized as follows:
\begin{tightlist} 
    \item We propose PrioriTouch, a framework that formulates contact preference learning as a ranking problem over control objectives and executes the learned priority ordering as a null space hierarchy via H-OSC for whole-arm pHRI.
    \item We introduce LinUCB‑Rank, a contextual bandit that learns priority orderings from sparse user feedback while accounting for inter‑objective coupling in hierarchical control.
    \item We enable safe and data-efficient learning through simulation-in-the-loop validation, where candidate priority updates are tested in a digital twin before real-world deployment. \looseness=-1
   \item We conduct a user study to inform realistic models of contact preferences for robot-initiated touch, which we leverage to simulate authentic user feedback in our evaluation.
   \item We evaluate PrioriTouch through extensive simulation, real-world experiments, and a realistic caregiving user study, demonstrating effective adaptation to individual contact preferences without compromising task performance or comfort.
\end{tightlist}

Our framework integrates user contact preference learning with low-level control by parameterizing operational space control using the outputs of a learned ranking policy. This structured integration ensures that high-level feedback is directly translated into low-level force regulation and pose tracking, effectively bridging the gap between user preferences and robot control.

\section{Related Work}
\vspace{-5pt}

\textbf{Whole-Arm Manipulation}. Recent whole-arm manipulation (WAM) systems leverage distributed tactile skins to localize and regulate contact \cite{xu2024cushsense, si2023robotsweater, goncalves2022punyo, liu2022warm}. These capabilities have been applied to manipulating heavy or bulky objects \cite{gliesche2021geometry, goncalves2022punyo}, navigating clutter \cite{jain2013reaching}, and providing social touch \cite{block2021six}. In pHRI scenarios, however, explicitly managing multiple, potentially conflicting contact-force objectives remains challenging. Park et al.\ \cite{park2008robot} proposed a hierarchical multi-contact controller that optimizes all contact forces with equal weighting, but their formulation does not account for preference-related ordering or any dynamic prioritization. In contrast, we build on hierarchical control but dynamically prioritize contact points using user feedback, enabling robots to accommodate individualized comfort and safety preferences across different users and body regions.


\begin{figure*}[t!]
    \centering
    \includegraphics[width=\textwidth]{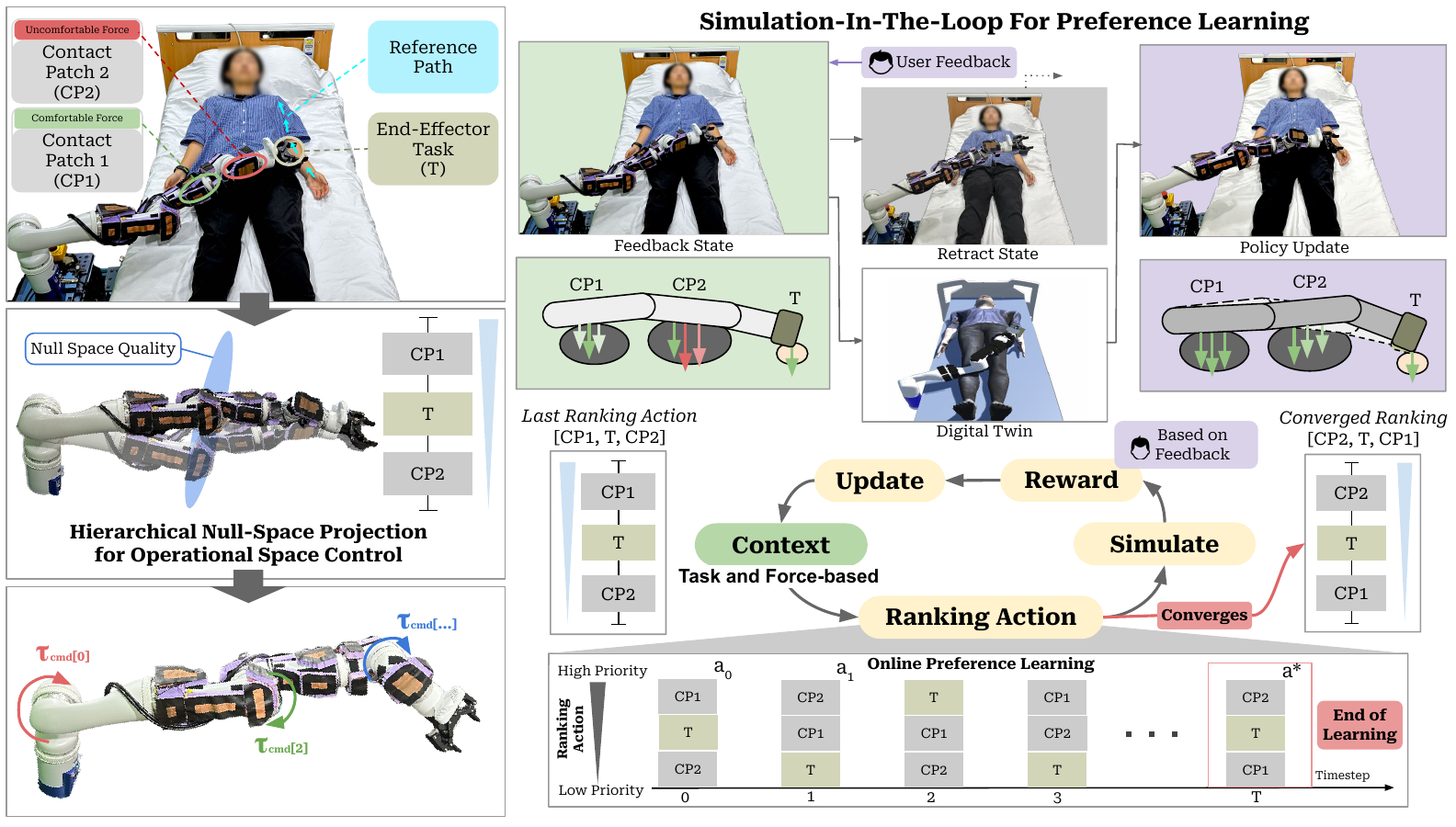}
    \vspace{-5pt}
    \caption{We implement H-OSC which uses a learned priority ranking to modulate multiple contacts during WAM. We propose LinUCB-Rank, a learning-to-rank contextual bandit algorithm, to update the prioritization policy using realistic user feedback and simulated interactions.}
    \label{fig:overview}
\end{figure*}

\textbf{Online Preference Learning in HRI.} While several works have explored online user preference learning in HRI, spanning high-level action sequences \cite{maroto2022preference} to low-level parameters \cite{losey2018including} and combinations thereof \cite{canal2019adapting, narcomey2024learning}, existing work has not explicitly addressed the challenge of learning user contact preferences. Most previous research either assumes human-initiated contact \cite{bajcsy2018learning, losey2019learning, losey2022physical, mehta2024strol} or focuses on user preferences that can be safely explored by adjusting non-contact parameters, such as trajectory speed \cite{canal2019adapting}. However, in pHRI tasks requiring robot-initiated physical contact, direct trial-and-error exploration to determine acceptable forces is impractical and potentially unsafe. Exploring contact preferences in multi-contact settings is especially challenging, as mitigating discomfort at one body region may inadvertently increase forces at another, making conflict resolution substantially harder. We address this gap by casting contact preference learning as a learning‑to‑rank problem over control objectives, allowing us to reason about preferences pertaining to multiple simultaneous contacts in a structured manner. We also introduce simulation-in-the-loop learning, allowing robots to refine candidate prioritization policies virtually before resuming real interaction, enabling safe and personalized estimation of user contact preferences in pHRI.

\section{Problem Statement}
\label{sec:problem-statement}
\vspace{-5pt}

\begingroup
\hyphenpenalty=100
\exhyphenpenalty=100
{
We ground the problem in whole-arm pHRI (e.g., bed bathing; Fig.~\ref{fig:overview}). An $n$-DoF robot follows a reference path
$\prefpath = \{\mathbf{p}(1),\ldots,\mathbf{p}(N)\}$,
where each $\mathbf{p}(i)\in\mathbb{R}^6$ is an end-effector pose.
At time $t$, the joint configuration is $\mathbf{q}_t$ and current end-effector pose is $\mathbf{p}_t^{\mathrm{curr}}$. The controller selects a waypoint index $i_t$ and defines the desired pose as $\mathbf{p}_t^{\mathrm{des}} \triangleq \mathbf{p}(i_t)$. The controller applies joint torques $\boldsymbol{\tau}_t\in\mathbb{R}^n$ to reduce the pose error and execute the task.

We model the human as a discrete set of body parts 
$\mathcal{B}=\{b_1,\ldots,b_m\}$ with pose configuration $\mathbf{h}_t$ at time $t$.
The contact set at time $t$ is $\mathcal{C}_t \;=\; \{\,c_t(1),\,c_t(2),\,\ldots,\,c_t(K_t)\,\}$,
where $K_t=|\mathcal{C}_t|$ is the number of contacts present at time $t$ and $c_t(k)$ denotes the $k$-th contact.
Each contact $c_t(k)\in\mathcal{C}_t$ is associated with a body part via
$\psi:\mathcal{C}_t\to\mathcal{B}$ and produces a force vector
$\mathbf{f}_t(c_t(k))\in\mathbb{R}^3$.
For a body part $b\in\mathcal{B}$, the set of contacts on $b$ is
given by $
\mathcal{C}_t(b) \;\triangleq\; \{\,c_t(k)\in\mathcal{C}_t \;\mid\; \psi(c_t(k))=b\,\},
$
and the aggregate force magnitude on $b$ is
$
f_t(b) \;\triangleq\; \max_{c_t(k)\in\mathcal{C}_t(b)} \big\|\mathbf{f}_t(c_t(k))\big\| ,
$
with $f_t(b)=0$ if $\mathcal{C}_t(b)=\varnothing$.
The set of body parts in contact at time $t$ is
$\mathcal{B}_t \triangleq \{\,b\in\mathcal{B}\mid \mathcal{C}_t(b)\neq\varnothing\,\}$.
For convenience, we jointly define the robot–human state as
$\mathsf{s}_t=(\mathbf{q}_t,\mathbf{h}_t,\mathcal{C}_t)$.

User contact preferences are represented by an unknown model $\usermodel$, which assigns a comfort threshold $\comfort{b}$ to each body part $b\in\skel$ (Sec.~\ref{par:study_preference}). Although $\usermodel$ is not directly observable, the robot receives sparse online feedback $\feedback{t}=\feedbackpair{b}{\delta}$, where $b\in\skel$ identifies the body part of concern, and $\delta$ is a one‑hot vector encoding the requested change. We fix the order as $\delta = [\texttt{dec-Large},\,\texttt{dec-Small},\,\texttt{inc-Small},\,\texttt{inc-Large}]$,
so a given $\delta$ selects decrease/increase and large/small. For example, “my stomach hurts a bit” implies $\feedbackpair{\texttt{abdomen}}{[0,1,0,0]}$,
i.e., a small decrease on abdomen. \looseness=-1

At each timestep $t$, the controller balances two objectives. The 
pose-tracking objective penalizes deviation from the desired waypoint, $J_{\mathbf{p}}(t) = \|\mathbf{p}^{\text{curr}}_t - \pdes{t}\|$. The force-regulation objective penalize comfort-threshold exceedances for each contacted body part, 
$J_{f,b}(t) = \max\!\bigl(0,\,\bodyforce{t}{b}-\comfort{b}\bigr)$ for $b\in\skel_t$. The active objective set is 
$\Jset{t}=\{J_{\mathbf{p}}(t)\}\cup\{J_{f,b}(t)\mid b\in\skel_t\}$.

Because objectives in $\Jset{t}$ may conflict depending on $\state$, they are executed according to a time-varying ordering 
$\sigma_t\in\mathfrak{S}(\Jset{t})$ (denotes all permutations of active objectives) using H-OSC 
(Sec.~\ref{subsec:hierarchical-osc}). A ranking policy $\pisig(t)$ predicts $\sigma_t$ at run time with the goal of tracking $\prefpath$ while respecting comfort thresholds with minimal user feedback. We formulate this as a contextual bandit (CB) problem. At each timestep $t$: the context
$\ctx{t}$ is derived from the observable state $\state$; the action is the ranking $\sigma_t$ drawn from the action space $\Ascr{}=\mathfrak{S}(\Jset{t})$; and the reward $\rew{t}\in\mathbb{R}$ is a scalar feedback signal returned after executing $\sigma_t$. The CB thus selects $\sigma_t$, and H-OSC executes 
it as a strict hierarchy. Details of how $\ctx{t}$ and $\rew{t}$ are constructed are provided in Sec.~\ref{subsec:linucb-rank}.
}
\endgroup

\section{PrioriTouch: Contact Preference Learning for Whole-arm Manipulation}
\label{sec:methods}
\vspace{-5pt}

\noindent PrioriTouch casts contact preference learning as a \emph{learning-to-rank} problem over control objectives. LinUCB-Rank, a contextual bandit, learns a priority ordering that H-OSC \cite{sentis2004prioritized} executes as a null space hierarchy, translating high-level contact preferences into low-level control. For safety and data efficiency, candidate re-orderings are validated in a digital twin before deployment. We now detail the components that enable this framework.

\subsection{Hierarchical Operational Space Control for WAM}
\label{subsec:hierarchical-osc}
\vspace{-5pt}
H-OSC~\cite{sentis2004prioritized} enables robots to manage multiple competing objectives by enforcing a strict hierarchy, making it particularly suitable for WAM scenarios with simultaneous pose tracking and force regulation. Given a ranking $\sigma_t$, H-OSC recursively projects lower-priority objectives into the null space of higher-priority
ones, ensuring that higher-priority tasks are executed without interference.

Formally, the robot dynamics can be written as
$\mathbf{M}(\mathbf{q}_t)\ddot{\mathbf{q}}_t + \mathbf{C}(\mathbf{q}_t,\dot{\mathbf{q}}_t) + \mathbf{g}(\mathbf{q}_t) = \boldsymbol{\tau}_t$. H-OSC generates torques as 
$\boldsymbol{\tau}_t = \sum_{j\in\Jset{t}} \mathbf{N}_{j,t}\,\boldsymbol{\tau}_{j,t}$, 
where $\boldsymbol{\tau}_{j,t} = \bar{\mathbf{J}}_{j}^\top \mathbf{F}_{j,t}$ and $\mathbf{N}_{j,t}$ is the recursively computed null space projector.

Operational-space forces $\mathbf{F}_{j,t}$ are derived from desired task accelerations. For pose tracking, we use $\ddot{\mathbf{x}}_t = \Kp(\pdes{t} - \mathbf{p}^{\text{curr}}_t) - \Kd\dot{\mathbf{p}}^\text{curr}_t$, where $\Kp$ and $\Kd$ are proportional and derivative pose gains. For force regulation at body part $b\in\skel_t$, we use $\ddot{\mathbf{x}}_t = -\Kf(\bodyforce{t}{b} - \comfort{b})$, where $\Kf$ is a scalar force gain. Detailed derivations appear in Appendix~\ref{subsec:appendix_hosc}.

\subsection{Learning to Rank using Contextual Bandits}
\label{subsec:linucb-rank}
\vspace{-5pt}

Enumerating and testing all permutations of objectives in $\Jset{t}$ is computationally infeasible. 
We therefore formulate control objective prioritization for H-OSC as a \emph{learning-to-rank} problem guided by user feedback. 
We use a \emph{contextual bandit} (CB)~\cite{bietti2021contextual} to balance exploration and exploitation, refining the ranking policy iteratively from sparse feedback.

\begin{wrapfigure}{r}{0.54\textwidth}           
  \vspace{-3ex}                                 
  \begin{minipage}{0.54\textwidth}
    \begin{algorithm}[H]
\caption{LinUCB-Rank}
\label{alg:linucb-rank}
\begin{algorithmic}[1]
\Require Objective set $\Jset{t}$, exploration parameter $\alpha > 0$
\State Initialize $(\hat{\boldsymbol{\theta}}_{i}, \mathbf{A}_{i}, \mathbf{b}_{i})$ for all $i = 1,\ldots,|\Jset{t}|$
\For{$t = 1$ to $T$}
    \State $\Jset{t}^{(0)} \gets \Jset{t}$,\; $\sigma_t \gets []$\;
    \State Initialize ranking history $\ctx{t,h}\gets \mathbf{0}$
    \For{$i = 1$ to $|\Jset{t}|$}
        \State Construct context $\ctx{t,i}$ from state $\state$
        \For{each $a_{t,i} \in \Jset{t}^{(i-1)}$}
            \State $\mathrm{UCB}_{a_{t,i}} \gets \hat{\boldsymbol{\theta}}_{i}^\top \ctx{t,i} + \alpha \sqrt{\ctx{t,i}^\top \mathbf{A}_{i}^{-1}\ctx{t,i}}$
        \EndFor
        \State $a_{t,i} \gets \arg\max_{a_{t,i} \in \Jset{t}^{(i-1)}} \mathrm{UCB}_{a_{t,i}}$
        \State $\sigma_t \gets \sigma_t + a_{t,i}$
        \State $\Jset{t}^{(i)} \gets \Jset{t}^{(i-1)} \setminus \{a_{t,i}\}$
    \EndFor
    \State Execute H-OSC with ranking $\sigma_t$
    \State Observe per-slot rewards $\{\rew{t,i}\}_{i=1}^{|\Jset{t}|}$
    \For{$i = 1$ to $|\Jset{t}|$}
        \State $\mathbf{A}_{i} \gets \mathbf{A}_{i} + \ctx{t,i}\ctx{t,i}^\top$
        \State $\mathbf{b}_{i} \gets \mathbf{b}_{i} + \rew{t,i}\ctx{t,i}$
        \State $\hat{\boldsymbol{\theta}}_{i} \gets \mathbf{A}_{i}^{-1}\mathbf{b}_{i}$
    \EndFor
\EndFor
\end{algorithmic}
\end{algorithm}
\end{minipage}
  \vspace{-4ex}                                
\end{wrapfigure}

\exhyphenpenalty=10000

\textbf{Preliminaries.} At each timestep $t$, a contextual bandit observes a context $\ctx{t}$, selects an action $\sigma_t \in \Ascr{t}$, and receives a reward $\rew{t}(\ctx{t},\sigma_t)$. The goal is to minimize the cumulative regret 
$\text{Regret}(T) = \sum_{t=1}^T \max_{\sigma \in \Ascr{t}} \mathbb{E}[\rew{t}(\ctx{t},\sigma)]$ $-$ $\sum_{t=1}^T \mathbb{E}[\rew{t}(\ctx{t},\sigma_t)]$ 
LinUCB~\cite{li2010contextual,gordon2020adaptive} assumes a linear reward model $\mathbb{E}[\rew{t} \mid \ctx{t},a] = \boldsymbol{\theta}_a^\top \ctx{t}$ for each atomic action $a$, and selects actions using an upper confidence bound (UCB): 
$\mathrm{UCB}_{t,a} = \hat{\boldsymbol{\theta}}_a^\top \ctx{t} + \alpha\sqrt{\ctx{t}^\top \mathbf{A}_a^{-1}\ctx{t}}$, 
where $\mathbf{A}_a$ is a covariance matrix and $\alpha>0$ controls exploration.

In our setting, the action is a full ranking $\sigma_t$. Standard LinUCB would choose a single best element $a$ per round, which is insufficient for prioritizing multiple coupled objectives in H-OSC. LinUCB-Rank instead constructs $\sigma_t$ sequentially, explicitly accounting for interdependencies: assigning higher priority to one objective reduces the null space available for those below it.\looseness=-1

\textbf{Contextual Bandits for Ranking.}  
We propose \emph{LinUCB-Rank} (Alg.~\ref{alg:linucb-rank}), which extends LinUCB to learn a ranking policy $\pisig(t)$ by sequentially assigning positions within each timestep. At every timestep $t$, LinUCB-Rank builds a ranking $\sigma_t = (a_{t,1},\ldots,a_{t,|\Jset{t}|})$ by selecting, at each slot $i$, the objective $a_{t,i} \in \Jset{t}^{(i-1)}$ with the highest UCB among the remaining candidates. While Alg.~\ref{alg:linucb-rank} assumes a fixed set of objectives, $\Jset{t}$ may vary due to changing contacts $\contactset{t}$. We therefore rank a \emph{fixed surrogate set}, i.e., the set of body parts $\skel$; run LinUCB-Rank over $\skel$ to obtain an ordering ${\rho}_t$, then sort active objectives by this order.

The context $\ctx{t,i}$, shared across all candidates at rank $i$, is derived from $\state$ and includes three parts:  
(1) \textbf{Max force per body part}, $\ctx{t,f}\in\mathbb{R}^m$, with $[\ctx{t,f}]_b=\bodyforce{t}{b}$;  
(2) \textbf{Pose error}, $\ctx{t,p}=\|\mathbf{p}^{\text{curr}}_t-\pdes{t}\|$;  
(3) \textbf{Ranking state}, $\ctx{t,h}\in\mathbb{R}^m$, an indicator vector that encodes the positions already assigned to each body part (0 if unassigned).

After constructing $\sigma_t$, H-OSC executes it. The robot then observes sensor data and user feedback $\feedback{t}$, which together define per-slot rewards $\rew{t,i}$ for each element $a_{t,i}$. Each reward has two terms:  
(1) \textbf{Feedback alignment} ($\rew{\feedback{t},i}$), positive if $a_{t,i}$ matches the body part indicated in $\feedback{t}$ and penalized otherwise depending on rank deviation;  
(2) \textbf{Threshold violation} ($\rew{t,f,i}$), penalizing excess force as $\rew{t,f,i} = -w_{f}\max_{b\in\skel}\bigl(0,\,\bodyforce{t}{b}-\comfort{b}\bigr)$, where $w_{f}>0$.  
The total reward is $\rew{t,i} = \rew{\feedback{t},i} + \rew{t,f,i}$.

\hyphenpenalty=100
\exhyphenpenalty=100

\subsection{Simulation-in-the-Loop Preference Learning}
\label{subsec:sim-in-the-loop}
\vspace{-5pt}

At run time, H-OSC generates low-level control commands by executing an objective hierarchy produced by the ranking policy $\pi_{\sigma}(t)$. This policy is updated online by LinUCB-Rank from sparse user feedback. Because LinUCB-Rank is a contextual bandit, it must \emph{explore} alternative orderings to align with user preferences. Performing this exploration directly on the user can induce large transients: in H-OSC, promoting one objective reduces the null space available to the rest, potentially increasing forces elsewhere. 

To keep adaptation safe and data-efficient, PrioriTouch adopts a \emph{simulation-in-the-loop} approach, performing high-risk exploration in a digital twin and deploying the converged policy on the real system. Specifically, the robot performs the following steps:
\vspace{-2pt}
\begin{tcolorbox}[colback=blue!3!white, colframe=blue!60!black, title=Simulation-in-the-Loop Preference Learning, boxrule=0.5pt, arc=5pt, left=6pt, right=1pt, top=0.5pt, bottom=0.5pt]
\begin{enumerate}[leftmargin=-1.5em, itemsep=1pt, parsep=1pt, align=right]
    \item[\stepicon{1}] \textbf{Record feedback:} Receive user feedback $\feedback{t}$, update threshold $\comfort{b}$, and log state $\mathsf{s}_t$.
    \item[\stepicon{2}] \textbf{Retract safely:} Move the robot to a predefined safe configuration.
    \item[\stepicon{3}] \textbf{Simulate exploration:} Set digital twin at state $\mathsf{s}_t$.
    \item[\stepicon{4}] \textbf{Learn preferences:} Iteratively refine policy $\pi_\sigma(t)$ using LinUCB-Rank (Sec.~\ref{subsec:linucb-rank}).
    \item[\stepicon{5}] \textbf{Resume operation:} Upon convergence, resume real-world interaction with updated $\pi_\sigma(t)$.
\end{enumerate}
\end{tcolorbox}
\vspace{-2pt}

Exploration occurs off-body in the twin, reducing the need for real-world user interactions to personalize the hierarchy. Only orderings that perform well under the simulated reward signals are deployed, yielding safer and more data-efficient adaptation.

\section{Experiments}
\vspace{-5pt}

\hyphenpenalty=100 \tolerance=100

We evaluate \textit{PrioriTouch} across three settings: (1) \textbf{simulation} to selectively evaluate learning and force modulation effects, (2) a \textbf{hardware testbed} to assess it on complex trajectories in the real world, and (3) \textbf{user studies} to model contact preferences and evaluate PrioriTouch in a real-world pHRI scenario. We use RCareWorld~\cite{ye2022rcare} for simulation and a Kinova Gen3 arm with distributed tactile sensing for all experiments; full system details are in the Appendix \ref{subsec:appendix_sysdetails}.

\begin{figure*}[t]
    \centering
    \includegraphics[width=0.95\textwidth]{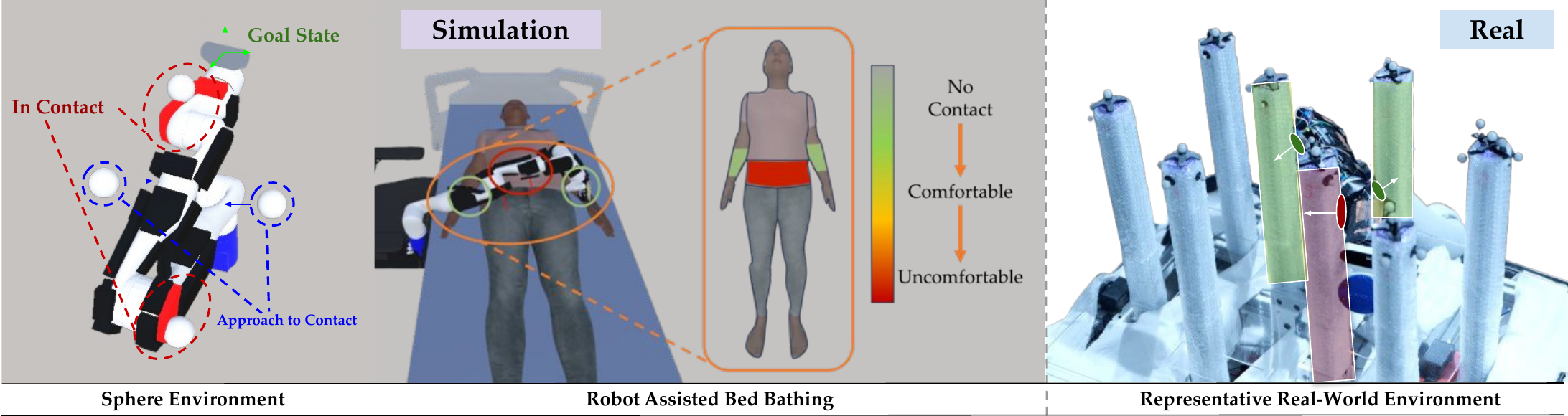}
    \vspace{-4pt}
    \caption{Simulation environment and Hardware testbed: (\emph{Left})~Sphere environment with multiple spheres colliding with the robot’s arm to test force modulation. (\emph{Middle})~Simulated robot-assisted bed-bathing scenario with whole-arm contact to wipe a user’s limbs. (\emph{Right})~Real-world 3D goal-reaching maze with vertical cylinders, each representing a unique body part with its associated contact preference.}
    \label{fig:exp_setup}
\end{figure*}

\noindent\textbf{Baselines}.
We compare LinUCB-Rank against the following methods:
\begin{tightlist}
\item \textbf{Fixed Priority (\emph{FP})}: Maintains a static priority ordering derived from aggregate user-study preferences (Sec.~\ref{par:study_preference}), without adapting based on feedback.
\item \textbf{Heuristic-based Priority (\emph{HP})}: Starts from the same initial ordering as \emph{FP} but reorders objectives based on violations of force thresholds. For multiple violations, contacts are prioritized based on the magnitude of the violations.
\item \textbf{LinUCB (Sorted)}: Independently learns contact force preferences using LinUCB, by sorting estimated upper confidence bounds.
This baseline ignores dependencies between objectives.
\end{tightlist}

\noindent\textbf{Evaluation Metrics}.\label{sec:metrics} We assess performance using:
\begin{tightlist}
\item \underline{Number of User Feedback Signals}: Number of $\feedback{t}$ received before achieving correct ordering. 
\item \underline{Force Threshold Violation}: Number of timesteps with $f_t(b) > \comfort{b}$.
\item \underline{Time to Task Completion}: Time taken to track the entire reference path $\prefpath$.
\item \underline{Average Force Exerted Per Body Part}: How well a method respects user force thresholds for different body parts across different paths.
\end{tightlist}

\paragraph{\noindent\textbf{Simulation Experiments\vspace{3pt}\newline}}

\hspace{-10pt}\noindent\textbf{(a) \textit{Sphere Environment.}} We simulate 2–4 spheres moving toward a stationary arm (Fig.~\ref{fig:exp_setup}, \emph{Left}) to isolate force regulation. Each sphere represents a body part with predefined priorities and comfort thresholds; feedback is simulated accordingly (see Appendix \ref{subsec:sim_feedback}). As shown in Fig.~\ref{fig:exp_results} (\emph{Top-Left}), \textit{LinUCB-Rank} attains better sample efficiency than \textit{LinUCB (Sorted)} as the number of spheres increases; with only two spheres, the simpler baseline is competitive, but the advantages of iterative ranking grow with multi-contact coupling.

\begin{figure*}[t!]
    \centering
    \includegraphics[width=0.95\textwidth]{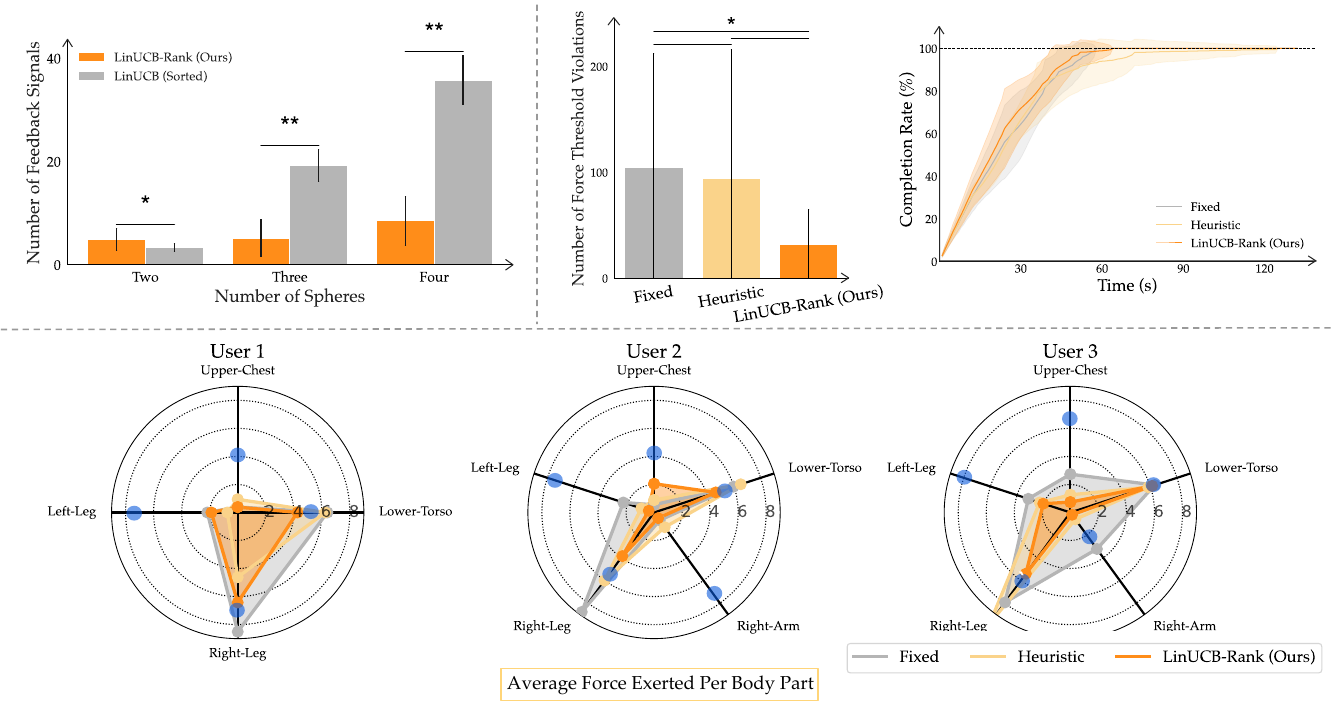}
    \vspace{-5pt}
    \caption{LinUCB-Rank requires fewer feedback signals (\emph{Top-Left}), fewer violations (\emph{Top-Right}), and obeys varying preferences related to force thresholds (denoted by a blue dot) across users (\emph{Bottom}). $\ast$ and $\ast\ast$ denote statistically significant differences with $p_{0.05}$ and $p_{0.005}$ respectively.}
    \label{fig:exp_results}
    \vspace{-3pt}
\end{figure*}

\noindent\textbf{(b) \textit{Robot-assisted Bed Bathing.}}
We simulate a wheelchair-mounted arm performing a sponge-wiping task on a user lying in a hospital bed (Fig.~\ref{fig:exp_setup}, \emph{Middle}). The path induces incidental whole-arm contacts with seven body regions. We evaluate \textit{LinUCB-Rank} against \emph{FP} and \emph{HP} over four trajectories and three user models with varied preferences. Results (Fig.~\ref{fig:exp_results}) show significantly fewer threshold violations than \emph{FP} and \emph{HP} (paired $t$-test, $p<0.05$). Unlike \emph{HP}, which reorders abruptly on violation spikes, \textit{LinUCB-Rank} anticipates preferences and reduces excessive forces. While \emph{FP} can finish faster by adhering to fixed priorities at the expense of comfort, \textit{LinUCB-Rank} maintains comparable completion times while improving comfort. It also adheres more closely to varying user thresholds (Fig.~\ref{fig:exp_results}, \emph{Bottom}).

\begin{figure}[h]
    \centering
    \includegraphics[width=0.96\textwidth]{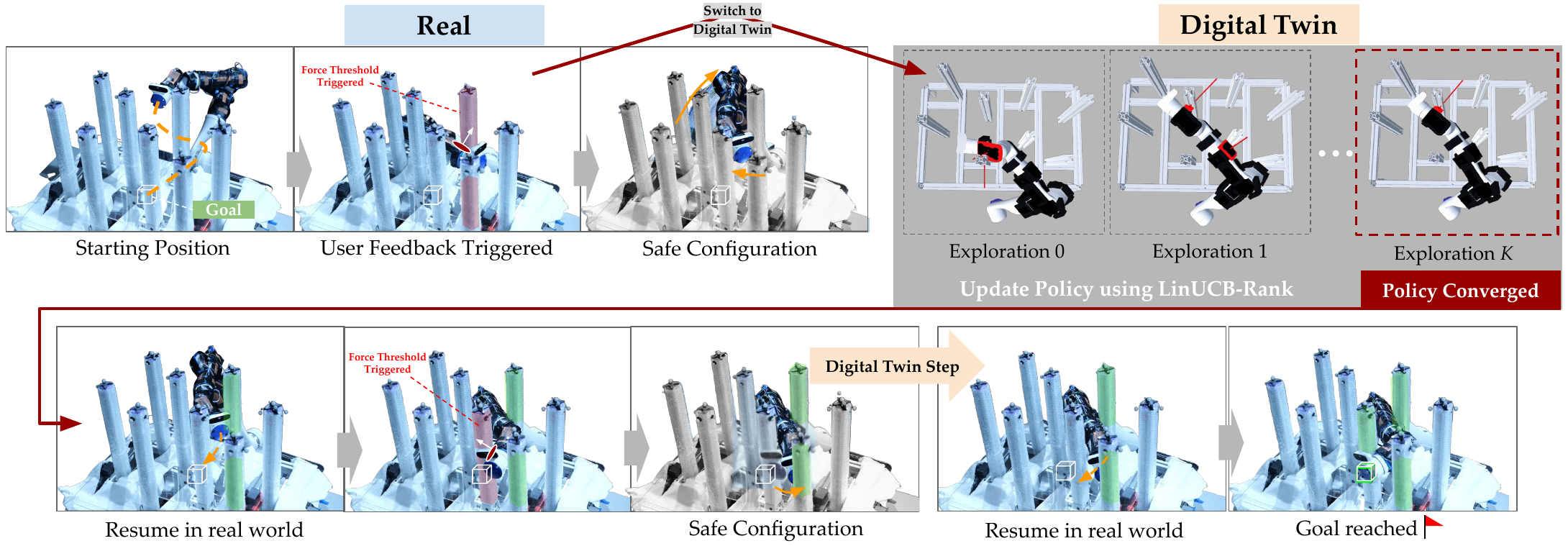}
    \vspace{-5pt}
    \caption{Real-world demonstration of PrioriTouch in a 3D goal-reaching maze. As the robot moves toward the goal, it periodically exceeds force thresholds (red cylinders), triggering user feedback. This feedback prompts LinUCB-Rank to update the priority policy within a digital twin. The robot then resumes with the refined policy, successfully progressing toward the goal.}
    \label{fig:real_demo}
\end{figure}

\paragraph{\noindent\textbf{Hardware Testbed.}}\label{par:testbed}
We deploy \textit{PrioriTouch} in a challenging real-world 3D goal-reaching maze (Fig.~\ref{fig:real_demo}), where vertical cylinders encode distinct body parts and preferences. The robot follows a predefined trajectory that necessarily induces whole-arm contact. When thresholds are exceeded, the robot retracts, updates the priority policy within a digital twin, and resumes with the refined policy, enabling steady progress toward the goal (see supplementary video on \href{https://emprise.cs.cornell.edu/prioritouch/}{website}).

\paragraph{\noindent\textbf{User Studies\vspace{3pt}\newline}}
\label{par:study_preference}
\hspace{-10pt}\noindent\textbf{(a) \textit{Eliciting Population-Level Contact Preferences.}}
We conducted a user study on robot-initiated contact in caregiving contexts (standing, bathing, dressing). Participants (n=98; ages 32–77, mean 58), recruited via Tetra Insights, included many with mobility limitations or disabilities. Each rated comfort for robot contact across 37 predefined body regions and three touch categories (functional assistance, emergency, sympathetic). Arms/hands and upper/lower back were most preferred for functional assistance, particularly among right-handed participants who needed support; sensitive regions (e.g., buttocks, genitals) were rarely selected. Fingers and toes were excluded because assistive tasks typically involve larger surface areas. See Appendix~\ref{subsec:cpref_study} for details.

\begin{wrapfigure}{r}{0.5\textwidth}
  \vspace{-12pt}
  \includegraphics[width=\linewidth]{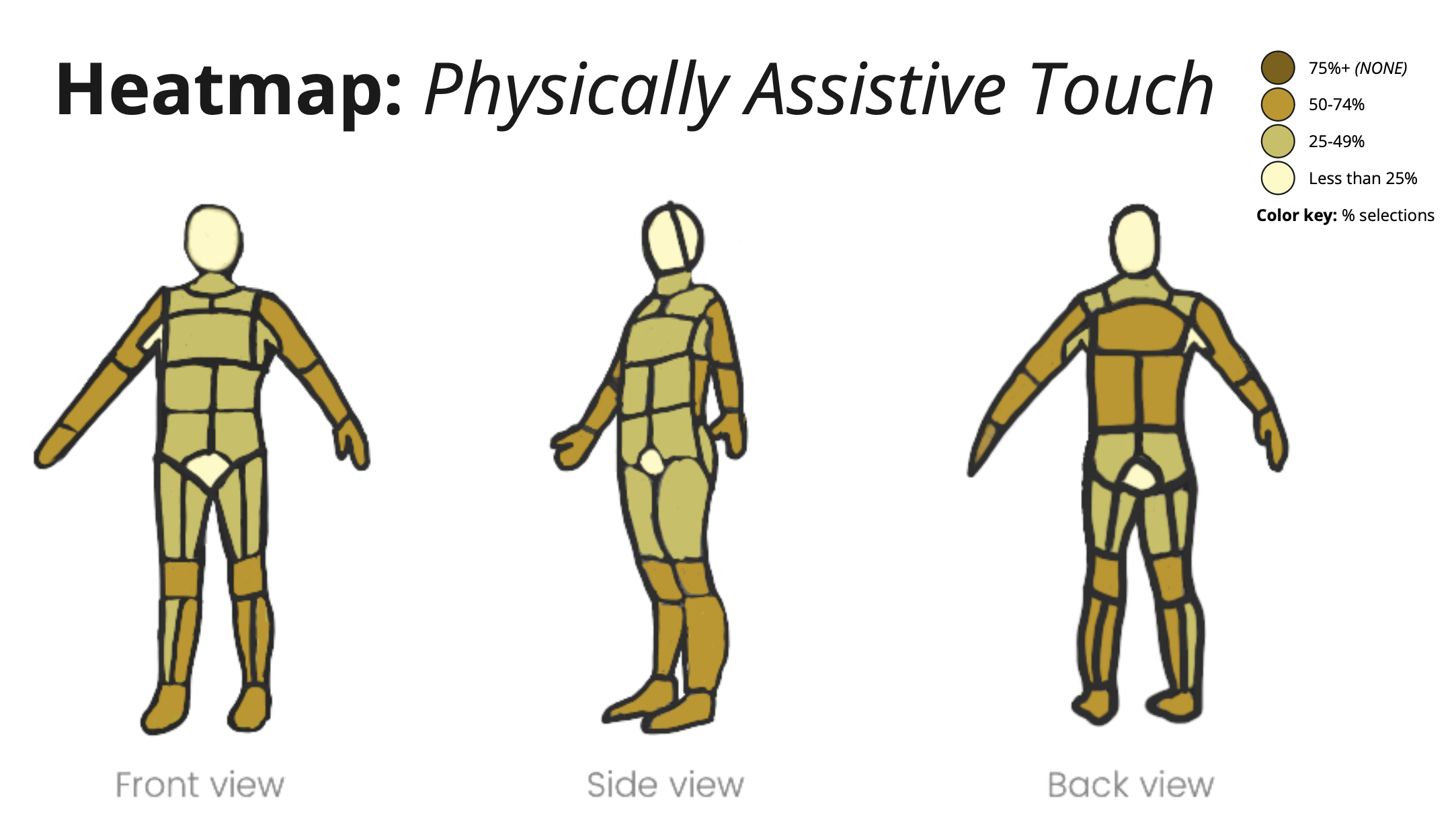}
  \vspace{-6pt}
  \caption{Outline of a human body divided into 37 regions; darker indicates greater touch acceptance, white indicates ``no touch.''}
  \label{fig:assistive-touch}
  \vspace{-4pt}
\end{wrapfigure}
\hyphenpenalty=5000 \tolerance=2000

\hyphenpenalty=100 \tolerance=100
These findings directly inform our realistic user modeling approach, which includes: 
(1) a \emph{base ranking} of contact openness by body part and 
(2) per-part \emph{comfort force thresholds}. At run time, the robot enforces these thresholds to avoid exceeding comfort limits and uses the base ranking to resolve concurrent violations. We combine survey-derived preferences with biomechanical pain limits (see Table~8 in~\cite{behrens2022statistical}) via $\comfort{b} = \gamma\,f^{\text{pain}}(b)\,S_r,$
where $f^{\text{pain}}(b)$ is from~\cite{behrens2022statistical}, $\gamma=0.7$ is a conservative base fraction, and $S_r$ is the sensitivity ratio computed as the selection frequency for a body part divided by the maximum frequency across parts. This normalization assigns higher ratios to frequently selected regions, yielding a realistic basis for modeling contact preferences.

\begin{wrapfigure}{r}{0.5\textwidth}  
  \vspace{-4pt}
  \includegraphics[width=\linewidth]
    {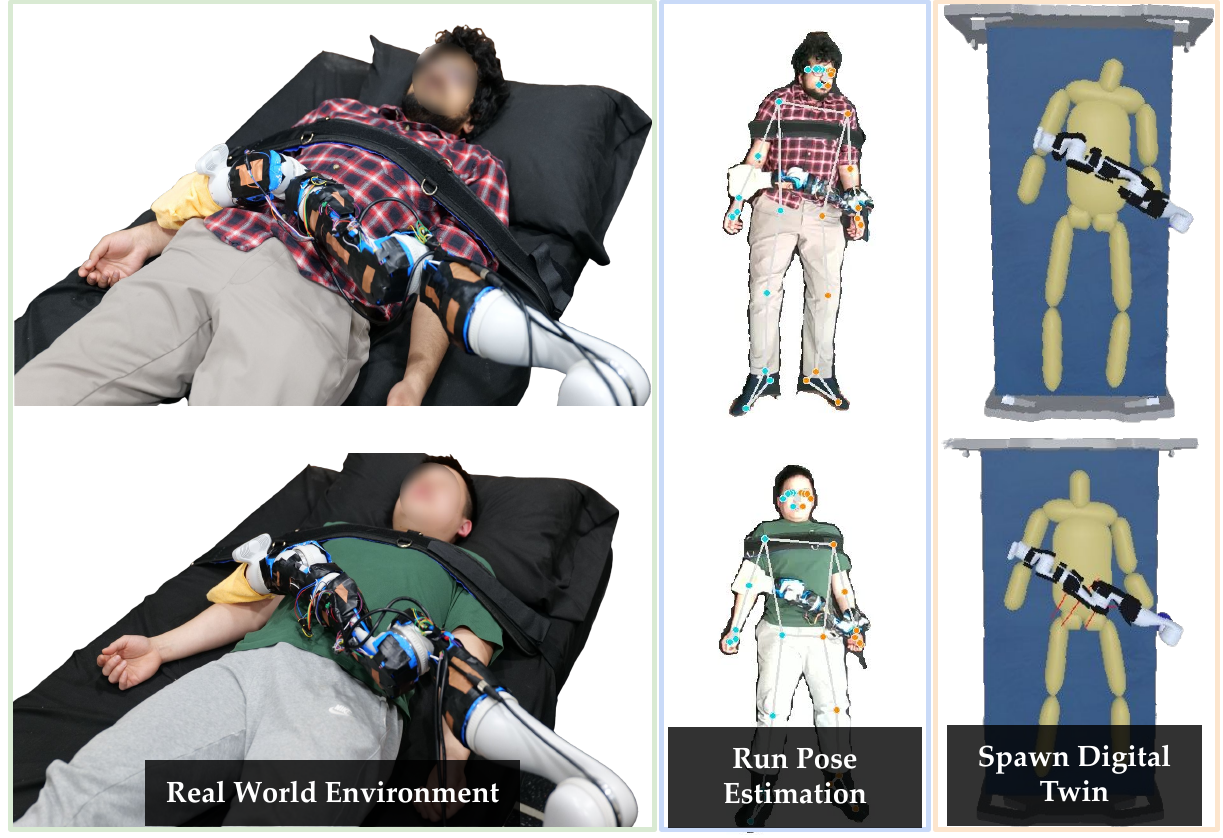}
  \caption{Experiment Setup: We use an overhead RGBD camera to perform pose estimation and spawn a digital twin for sim-in-the-loop preference learning.}
  \label{fig:real-study}
  \vspace{-3pt}
\end{wrapfigure}

\noindent\textbf{(b) \textit{PrioriTouch Evaluation with Humans.}}
\hyphenpenalty=5000 \tolerance=2000
To validate the practical feasibility of our approach in caregiving scenarios, we conducted a user study with 8 participants (without visible mobility limitations) performing the robot-assisted bed-bathing task (Fig. \ref{fig:real-study}). We use an RGBD camera (RealSense D455) and OpenPose to generate an aligned digital twin. In this evaluation, 7 out of 8 participants preferred our approach over the baseline (\emph{HP}) for tasks involving WAM. Our method received higher ratings in perceived safety, comfort, and overall task performance (see Appendix \ref{subsec:real_study}). Qualitative feedback was strongly positive, including comments such as: ``\textit{Both methods initially got stuck, but [ours] did pretty well on the second attempt}'' (second attempt implies post policy update), and ``\textit{It performed the task very well, the contacts felt seamless and natural.''}

\section{Limitations}
\label{sec:limitations}
\vspace{-5pt}

Although PrioriTouch provides a novel way of integrating user contact preferences into low-level control for whole-arm manipulation in pHRI, it has several limitations.

\noindent\textbf{Tactile Sensing and digital twin fidelity.}
PrioriTouch benefits from distributed tactile sensing and a digital twin for simulation-in-the-loop learning. While our experiments indicate robustness to moderate modeling errors in collision modeling (see Appendix \ref{subsec:additional_experiments}), the use of more accurate tactile sensing and richer human avatars, obtained using real-time tactile and visual feedback \cite{swann2024touch, chu2025humanrig}, would further improve sensitivity and sim-to-real transfer.

\noindent\textbf{Handling ambiguous or noisy user feedback.}
Contact preference learning is currently driven by sparse real-time verbal feedback parsed into structured signals. Highly ambiguous early feedback can slow personalization; uncertainty-aware querying and complementary cues (e.g., visuo–tactile patterns) are promising extensions.

\noindent\textbf{Access to reference path with whole-arm contact.}
We currently assume the robot has access to a feasible reference path that induces whole-arm contact, which is produced via teleoperation in our current workflow. Future work can explore contact-aware path planning, allowing \textit{PrioriTouch} to support a broader range of assistive tasks (e.g., transferring, dressing) requiring more dexterous whole-arm maneuvers.

\noindent\textbf{Real-world deployment.}
Lastly, we intend to deploy our method to assist individuals with mobility limitations. This will require unifying the improvements outlined above, including higher-fidelity sensing, better digital twins, and contact-aware planning, into a reliable whole-arm pHRI framework suitable for real-world use.

\acknowledgments{This work was partly funded by Toyota
Research Institute (TRI), National Science Foundation IIS \#2132846, and CAREER \#2238792. This article solely reflects the opinions and conclusions of its authors and not TRI or any other Toyota entity.}


\clearpage


\bibliography{references}  

\clearpage
\appendix
\section{Appendix}

\subsection{Detailed Derivation of Hierarchical Operational Space Control (H-OSC)}
\label{subsec:appendix_hosc}

H-OSC executes multiple objectives with a strict priority ordering. 
Consider an $n$-DOF robot manipulator with dynamics
\[
    \mathbf{M}(\mathbf{q}_t)\ddot{\mathbf{q}}_t + \mathbf{C}(\mathbf{q}_t,\dot{\mathbf{q}}_t) + \mathbf{g}(\mathbf{q}_t) = \boldsymbol{\tau}_t,
\]
where $\mathbf{q}_t \in \mathbb{R}^n$ is the joint configuration, 
$\mathbf{M}(\mathbf{q}_t)\in \mathbb{R}^{n\times n}$ is the inertia matrix, 
$\mathbf{C}(\mathbf{q}_t,\dot{\mathbf{q}}_t)\in \mathbb{R}^n$ are Coriolis/centrifugal terms, 
and $\mathbf{g}(\mathbf{q}_t)\in\mathbb{R}^n$ are gravity torques.

Given a priority ranking $\sigma_t=(a_{t,1},\dots,a_{t,|\Jset{t}|})$ over the active objectives $\Jset{t}$, 
H-OSC computes torques by projecting lower-priority objectives into the null space of higher-priority ones:
\[
    \boldsymbol{\tau}_t = \sum_{j\in\Jset{t}} \mathbf{N}_{j,t}\,\boldsymbol{\tau}_{j,t},
\]
with each unprojected torque 
$\boldsymbol{\tau}_{j,t} = \bar{\mathbf{J}}_{j,t}^\top \mathbf{F}_{j,t}$, 
where $\bar{\mathbf{J}}_{j,t}$ is the null space projected Jacobian. 
The null space projectors are updated recursively as
\[
    \mathbf{N}_{j+1,t} = \mathbf{N}_{j,t} - \mathbf{J}_{j,t}^\#\bar{\mathbf{J}}_{j,t}\mathbf{N}_{j,t}, 
    \quad \mathbf{N}_{1,t} = \mathbf{I}.
\]
Here $\mathbf{J}_{j,t}$ is the Jacobian of objective $j$, and $\mathbf{J}_{j,t}^\#$ its dynamically consistent generalized inverse:
\[
    \mathbf{J}_{j,t}^\# = \mathbf{M}^{-1}\bar{\mathbf{J}}_{j,t}^\top \left(\bar{\mathbf{J}}_{j,t}\mathbf{M}^{-1}\bar{\mathbf{J}}_{j,t}^\top\right)^{-1}.
\]

The operational-space dynamics for objective $j$ at time $t$ are
\[
    \mathbf{\Lambda}_{j,t}\ddot{\mathbf{x}}_{j,t} + \mathbf{\mu}_{j,t} + \mathbf{g}_{j,t} = \mathbf{F}_{j,t},
\]
where $\mathbf{\Lambda}_{j,t}$ is the operational-space inertia, 
$\mathbf{\mu}_{j,t}$ the Coriolis/centrifugal effects, 
and $\mathbf{g}_{j,t}$ gravity terms.

Desired task accelerations $\ddot{\mathbf{x}}_{j,t}$ depend on the objective type:
\begin{itemize}
\item Pose-tracking objective: $J_{\mathbf{p}}(t)$:  
  $\ddot{\mathbf{x}}_t = \Kp(\pdes{t} - \mathbf{p}^{\text{curr}}_t) - \Kd\dot{\mathbf{p}}^{\text{curr}}_t,$
  where $\Kp,\Kd$ are pose gains.
\item Force-regulation objective: $J_{\mathbf{f},b}(t)$ for body part $b\in\skel_t$:  
  $\ddot{\mathbf{x}}_t = -\Kf\,(\bodyforce{t}{b} - \comfort{b}),$
  where $\bodyforce{t}{b}$ is the measured aggregate force and $\comfort{b}$ its comfort threshold.
\end{itemize}

Finally, operational-space forces $\mathbf{F}_{j,t}$ are mapped back into joint torques as 
$\boldsymbol{\tau}_{j,t} = \bar{\mathbf{J}}_{j,t}^\top\mathbf{F}_{j,t}$ 
and then combined using null space projectors to enforce the ranking $\sigma_t$.
\subsection{Simulating User Feedback}
\label{subsec:sim_feedback}

We create a user model by initializing a base priority ID and comfort threshold for each body part. Given contact forces at timestep $t$, we compare the forces with the corresponding force thresholds. In case there is a body part for which the forces are above the threshold and the degree by which the forces are off. Using a sensitivity ratio (see Sec. \ref{par:study_preference}), which is a function of the comfort threshold itself, we generate the feedback for this body part. In case of multiple violations, we use the base priority ID to decide which body part to generate feedback for. Feedback is generated for body parts with higher base priority unless the force violation is beyond a safety force limit (set manually and the same for each user).
\label{sec:appendix}

\subsection{Comparison of Feedback Mechanisms Using NASA-TLX}
\label{sec:exp_user_study}

Our system currently solicits user feedback in a \emph{descriptive} form, where users describe the affected body part and how they feel (e.g., “I feel slightly uncomfortable around my abdomen”). An alternative is \emph{magnitude-based} feedback, in which users specify a numerical change in force (in Newtons). While magnitude-based feedback offers more precision, we hypothesize that most users prefer providing descriptive input due to the lower mental burden of recalling numeric force values. To investigate this, we conducted a Wizard-of-Oz study comparing these two feedback mechanisms via a modified NASA-TLX workload assessment.

\begin{figure*}
    \centering
    \includegraphics[width=\textwidth]{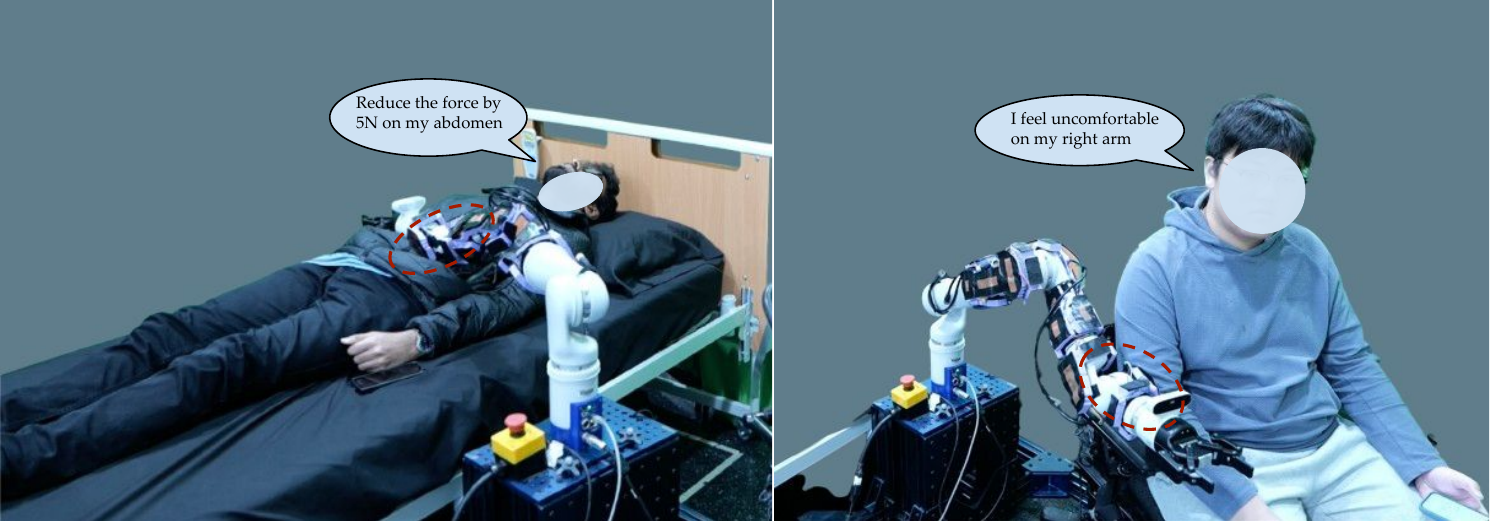}
    \caption{User study setup for comparing different feedback mechanisms in two different configurations. \emph{Left} shows the participant lying down on the bed along with an example of the magnitude-based feedback, and \emph{Right} shows the participant sitting on a wheelchair with an example of the descriptive feedback.}
    \label{fig:study_feedback}
\end{figure*}

\noindent\textbf{Study Procedure.}

We recruited 11 participants (8 male, 3 female; ages 21--28) without mobility limitations. Each participant experienced two configurations (see Figure \ref{fig:study_feedback}): i) sitting in a wheelchair with the robot’s forearm contacting their right arm, and ii) lying on a bed with the robot’s forearm or upper arm contacting their lower abdomen. For each configuration, participants performed a separate trial under each of the two feedback mechanisms. At the end of every trial, they rated the method on a 5-point Likert scale for mental demand, hurriedness, and irritation. We repeated each configuration two times, counterbalancing the order of feedback mechanisms across participants. This study was approved by our organization’s Institutional Review Board, and all participants provided written consent.

\begin{figure}[h!]
    \centering
    \includegraphics[width=0.6\columnwidth]{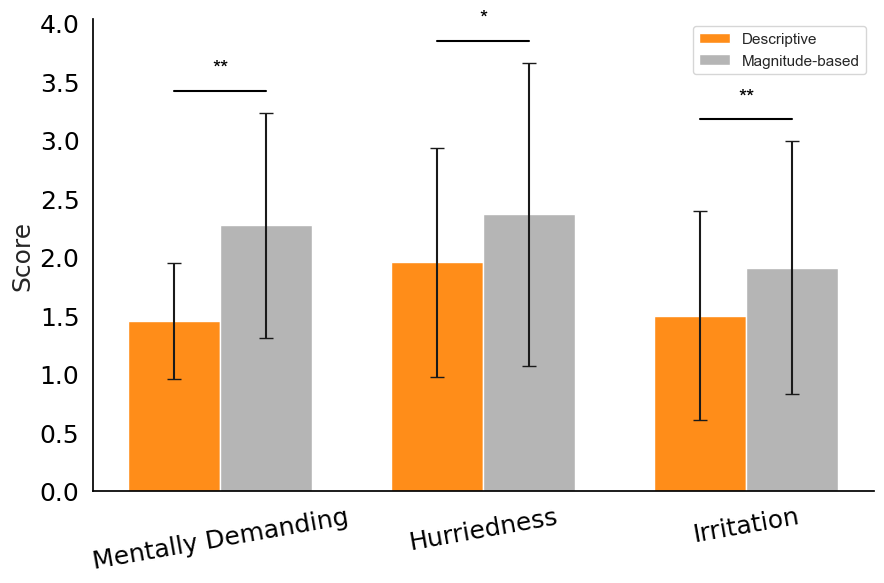}
    \caption{User study with 11 participants. Description-based feedback yields significantly lower cognitive workload (lower score is better) than magnitude-based controls.}
    \label{fig:subfig2}
\end{figure}

\noindent\textbf{Results and Analysis.}
Ten of the 11 participants preferred description-based feedback. Figure~\ref{fig:subfig2} shows that description-based feedback yielded significantly lower cognitive workload than the magnitude-based approach. A paired-sample \emph{t}-test indicated statistically significant differences in mental demand ($p<0.001$), hurriedness ($p<0.05$), and irritation ($p<0.001$) between the two methods. These findings support our hypothesis that descriptive feedback results in significantly lower mental burden, presenting a reasonable tradeoff between ease of feedback and precision of feedback.

\subsection{Additional System Details}
\label{subsec:appendix_sysdetails}
Sensors are modeled as soft bodies in Obi (Unity) and made using piezoresistive taxels (modified from \cite{xu2024cushsense}) in the real-world implementation. Each sensor provides contact location and force data. We integrate these sensor readings with H-OSC implemented using PyBullet~\cite{coumans2021}. H-OSC with LinUCB-Rank can operate at up to approximately 250 Hz, interfacing with a compliant low-level controller (1 kHz), ensuring robust handling of abrupt motions, such as muscle spasms. After stabilization, the system quickly refocuses on reference trajectories.

\subsection{Additional Details of Contact Preference User Study}
\label{subsec:cpref_study}

\textbf{Study Procedure.}
Participants were recruited via Tetra Insights, a platform providing financially compensated, high-quality respondents. Our study was approved by our Institutional Review Board. Each participant was asked to envision a home-helper robot with a human-like form factor depicted through a hand-sketched illustration featuring inflated, soft materials. Participants indicated their comfort with robot-initiated touch on a graphical interface that included 37 predefined body regions, identified with input from two medical doctors (general practitioners). Participants also had the option to provide additional insights through open-ended responses.

\textbf{Participant Demographics.}
The study included 98 adults (ages 32–77, mean age 58), of whom 42 participants were aged 65 or older and 56 were younger. Among participants, 67 identified as female (including 1 transgender female), and 31 identified as male. Forty-nine participants reported having a disability, 42 reported past injuries, and 7 chose not to disclose.

\textbf{Detailed Findings.}
Overall, 40\% of participants provided detailed open-ended responses elaborating on their choices. Arms and hands received the highest number of selections (779 total) across all touch categories. Right-handed participants specifically favored their dominant arm or hand for lighter supportive tasks, whereas the upper and lower back regions were predominantly selected for scenarios involving more substantial support or bracing. Regions such as the buttocks and genitals received minimal selections across all categories, highlighting privacy and comfort boundaries. Fingers and toes were deliberately excluded, as assistive interactions typically involve broader body regions.

\subsection{Real-world user study with human subjects}
\label{subsec:real_study}
\textbf{Study Procedure}. Participants first received an introduction describing the purpose of the study—evaluating two robotic arm control methods designed for robot-assisted caregiving involving whole-arm contact. Participants were informed about the specific caregiving scenario (bed bathing), their evaluation tasks (performance in clearing artificial dirt (ground coffee), perceived safety, and comfort), and how to provide verbal feedback during uncomfortable contacts. Before experimental trials, participants completed a supervised sample trial to practice providing feedback and simulating a person with mobility limitations. They used a provided digital questionnaire to record evaluations after each trial, identified via method IDs. In the actual trials, each participant experienced two distinct contact configurations across two methods, with two trajectories each (wiping arm and leg). Participants were repeatedly reminded to keep their responses independent of previous trials and to remain stationary during contact interactions. After completing all trials, participants filled out a final questionnaire before concluding the study.

\textbf{Results}. Below are the results from the study conducted with 8 participants with no visible mobility limitations.
\begin{table}[ht]
  \centering
  \caption{User–study ratings (mean ± SD on a 1–5 Likert scale).}
  \label{tab:ratings}
  \begin{tabular}{l
                  S[table-format=1.2(2)]
                  S[table-format=1.2(2)]
                  S[table-format=1.2(2)]}
    \toprule
    & {\bfseries Task Perf.} & {\bfseries Safety} & {\bfseries Comfort} \\
    \midrule
    Heuristic &
      2.53 \pm 0.94 &
      4.59 \pm 0.71 &
      4.65 \pm 0.61 \\[2pt]
    LinUCB-Rank (Ours) &
      {\textbf{$3.06 \pm 0.97$}} &
      {\bfseries $4.65 \pm 0.61$} &
      {\bfseries $4.82 \pm 0.39$} \\
    \bottomrule
  \end{tabular}
\end{table}

In the post-study questionnaire, participants were asked to select which of the two methods they would prefer for caregiving applications involving whole-arm contact. Seven out of eight participants preferred our method.

\subsection{Additional Experiments} \label{subsec:additional_experiments}

\noindent\textbf{Simulation-in-the-Loop Learning: Pauses and Accuracy.}

\noindent\textbf{Pauses:} We measured execution times for preference learning convergence in simulation ($52.4\pm 21.7$s). These pauses reduced progressively (to $\sim$30s) as user-contact preference estimates improved, indicating suitability for online use.

\noindent\textbf{Simulation Fidelity (Sim-to-Real Gap):} To test robustness to simulation inaccuracies, we varied the fidelity of the digital twin’s collision model (quality = 1: exact fit to visual mesh; quality = 0.75: 75\% fit). Lower fidelity led to only a modest (12.21\%) increase in required feedback signals, demonstrating our method’s resilience.

\vspace{1ex} \noindent\textbf{Comparison with Alternative Baselines.}

We evaluated PrioriTouch against two additional baselines: \begin{tightlist} \item \textit{Residual Reinforcement Learning:} Did not achieve stable convergence within our online-learning constraints, showing no force-violation reduction even after 20 minutes of training. \item \textit{Jain et al.’s MPC-based whole-arm manipulation (IJRR 2013):} Adapted to our trajectories, this baseline resulted in roughly 3x more force-threshold violations than PrioriTouch. \end{tightlist}

\end{document}